\documentclass{article}

\usepackage{microtype}
\usepackage{graphicx}
\usepackage{color,soul}
\usepackage{subfigure}
\usepackage{booktabs} % for professional tables
\usepackage{algorithm,amsmath}
\usepackage[noend]{algorithmic}
\usepackage{tabularx}
\usepackage{hyperref}

\usepackage[all=normal,floats,leading]{savetrees}
\usepackage[accepted]{mlsys2022}
\mlsystitlerunning{Bit-serial Weight Pools}

\begin{document}

\twocolumn[
\mlsystitle{Bit-serial Weight Pools: Compression and Arbitrary Precision Execution of Neural Networks on Resource Constrained Processors}

% unless you've provided the [accepted] option to the mlsys2022


\begin{mlsysauthorlist}
\mlsysauthor{Shurui Li}{ucla}
\mlsysauthor{Puneet Gupta}{ucla}
\end{mlsysauthorlist}

\mlsysaffiliation{ucla}{Department of Electrical and Computer Engineering, University of California, Los Angeles, California, USA}
\mlsyscorrespondingauthor{Shurui Li}{shuruili@ucla.edu}

\mlsyskeywords{Machine Learning, MLSys}

\vskip 0.3in

\begin{abstract}
Applications of neural networks on edge systems have proliferated in recent years but the ever increasing model size makes neural networks not able to deploy on resource-constrained microcontrollers efficiently. We propose bit-serial weight pools, an end-to-end framework that includes network compression and acceleration of arbitrary sub-byte precision. The framework can achieve up to $8\times$ compression compared to 8-bit networks by sharing a pool of weights across the entire network. We further propose a bit-serial lookup based software implementation that allows runtime-bitwidth tradeoff and is able to achieve more than $2.8\times$ speedup and $7.5\times$ storage compression compared to 8-bit weight pool networks, with less than $1\%$ accuracy drop.
\end{abstract}
]

\printAffiliationsAndNotice{}
\section{Introduction}
 The ever increasing size of neural network models and rapid proliferation of machine learning in resource-constrained edge devices have catalyzed research into a variety of model compression techniques, as well as software and hardware acceleration of deep learning on edge devices. 
 
 General-purpose microcontrollers have been a platform of choice for edge devices due to their low power, low cost and programmability. However, this comes at the cost of limited memory: these processors usually do not have any DRAM and often have less than 2MB total memory (SRAM + Flash); and small available compute power: these processors usually have small datapaths and  simple pipelines running at modest clock rates. This makes the execution of complex machine learning models on this ubiquitous class of processors very challenging. A variety of model compression techniques have, therefore, garnered attention in the embedded machine learning community \cite{compressionsurvey_tinyml}. 

Weight sharing \cite{softweightsharing1992} as a model compression technique shares a set of weight vectors across the entire neural network, so that only the indices of the shared weight vectors need to be stored, instead of actual weight values. For convolutional neural networks (CNNs), weight sharing methods can achieve compression ratios between 4-16x, compared to 8-bit baselines. Since weight sharing does not modify the structure nor the precision of the network, it can be combined with other compression techniques like pruning and quantization to further improve compression ratio and runtime. Furthermore, recent works \cite{choi2018pact_quantization,4bitquantization} have shown that sub-byte quantization of weights and/or activations can achieve inference accuracies comparable to full-precision networks.

Though weight sharing and sub-byte quantization are both promising for storage and runtime improvement, neither has native support in microcontroller-class general purpose processors commonly deployed in edge devices. As a result, these compression techniques can often hurt performance rather than improve it. For instance, processing a neural network with sub-byte precision naively can lead to worse runtime due to bit unpacking overhead \cite{Hu2018bitflow}. Hence, there is a need for optimized software implementations of weight-shared neural networks, as well as methods that can support and accelerate sub-byte precision neural networks on microcontrollers.

In this work, we present a framework for efficiently deploying large neural networks on small microcontrollers. The proposed framework contains two parts. The first part is neural network compression, where a pool of weight vectors (e.g., a $1\times 8$ 8-tuple of weights) along channel dimension are shared across the entire network. We refer to networks using our weight sharing method as weight pool networks in the rest of this paper. The second part of the framework is the software implementation of weight pool networks on microcontrollers, where we utilize bit-serial lookup tables to support and accelerate weight pool networks with 8-bit or lower activation bitwidth. 
%The proposed frame work uses weight sharing for net work compression, and utilizes bit-serial lookup tables to support and accelerate sub-byte activation bitwidths.
The main contributions can be summarized as follows.
\begin{itemize}
    \item We show that $z$ dimension weight pools, as small as 512 total parameters can realize popular networks such as ResNet and MobileNet with negligible accuracy loss.
    \item We develop a bit-serial lookup based method for efficient arbitrary-precision execution of weight pool networks on general purpose microcontrollers. This delivers 2.38X speedup (compared to well-optimized ARM CMSIS-NN library \cite{lai2018cmsis}) at 8-bit precision and even greater speedup at lower bitwidth on popular neural networks.
    \item   We explore the design-space of weight pool networks experimentally to develop an optimized software implementation of weight pool networks targeted for small, memory-starved microcontrollers.
    \item We show that weight pool arbitrary precision networks can be 2.8X faster and 6.51X more compact than CMSIS on ResNet-10, with less than 1 percent drop of accuracy on CIFAR-10, and better compression and speedup can be achieved on larger networks.
\end{itemize}

Next section outlines the motivation behind the bit-serial weight pool approach.

\section{Addressing Compression and Quantization Challenges for General Purpose Processors}\label{moitvation}

\paragraph{Compression with weight pools.}

Our weight pool networks essentially store vectors of weights along the channel dimension as one entry. The 3D filters used in CNNs would then be composed of these vectors. For instance, a $3 \times 3 \times 32$ filter would use $3 \times 3 \times 4 (=36) 1\times 8$ weight vectors selected from the available  pool of weight vectors. There is no limitation on the reuse of vectors. Weight-pool networks would reduce the parameter storage from the total number of parameters in the network to the total size of the weight pool. If done correctly, this can reduce parameter storage requirements of neural networks by orders of magnitude with minimal  accuracy drop. Furthermore, the parameter storage here becomes independent of network size. 

However, naively implementing weight pool networks would likely worsen inference latency due to additional memory reads (some form of index storage lookup followed by the actual weight lookup) with no reduction in total compute operations. One could try reducing the number of operations by directly storing the results of the (partial) dot product on the weight pools. For a pool vector size of 8, it would replace 8 multiply-accumulate operations with one memory lookup. Unfortunately, for 8-bit activations, this would require a lookup table size of ${2^8}^8$ entries for just one pool vector which is impractical.

\paragraph{Arbitrary precision computation using bit-serial arithmetic.}
Like conventional neural networks, the activation bitwidth of weight pool networks can be reduced to sub-byte regions while still achieving decent accuracy on many tasks. The sub-byte activation bitwidth provides an opportunity to improve the runtime and overall energy efficiency.

Sub-byte precision is not well supported in most microcontrollers (or most processors in general). Naively implementing networks with sub-byte activation bitwidth is not useful as it would worsen runtime due to the bit unpacking overhead with no actual compute reduction (since underlying hardware still executes higher precision arithmetic).  

To support and accelerate neural networks with sub-byte activation bitwidth, bit-serial multiplication  seems to be a suitable candidate since it processes a multiplication serially by looping through all the bits of one operand. The runtime of bit-serial multiplication is proportional to the bitwidth of the bit-unrolled operand. There are many bit-serial multiplication based hardware neural network accelerators \cite{li2021swis, judd2016stripes, sharma2018bitfusion}, but bit-serial multiplication is not supported in microcontrollers due to lack of bit-serial multipliers. 

\paragraph{Bit-serial-lookup-based weight pool networks.}
We address the challenges outlined above by doing bit-serial execution {\em but} saving computation by lookup of partial dot product results on pool vectors. Since activations are processed one bit at a time (most significant to least significant bit), the dot product lookups only need to be on 1-bit operands. Therefore, the lookup table for activation bitwidth of 8 bits is just $2^8$ entries. This would replace 8 multiply-accumulate operations with 8 memory reads and accumulations. Later we show how despite this, substantial runtime reduction can be achieved by careful implementation optimizations leveraging the value reuse properties of weight pools. Furthermore, reducing activation bitwidth now just amounts to truncating the temporal bit-serial execution earlier which gives proportionate further runtime improvement. 

%Next section describes our bit-serial weight pool approach in detail. 

\section{Bit-Serial Weight Pool Methodology}
\begin{figure}[h]
    \centering
    \includegraphics[width=1.0\columnwidth]{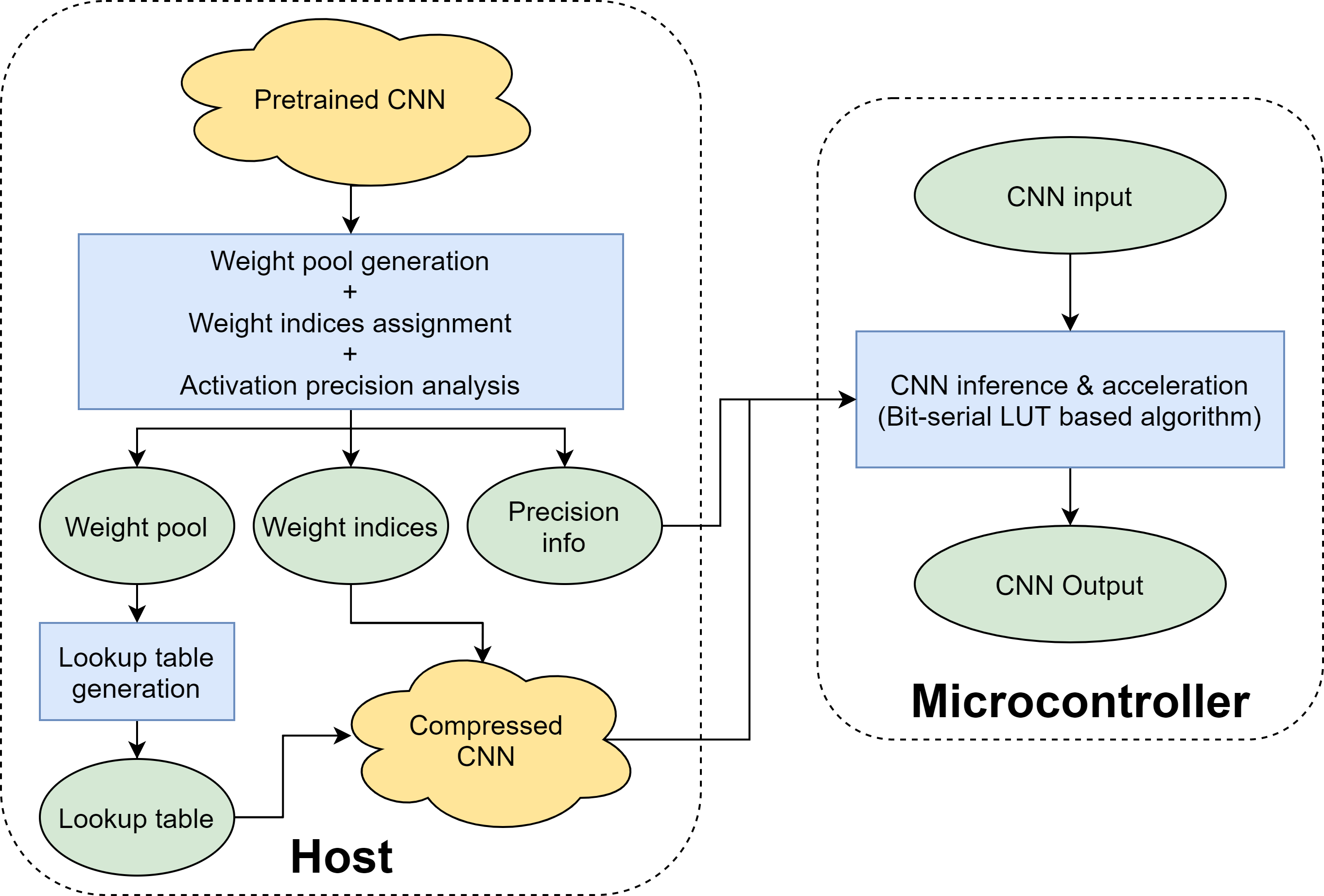}
    \caption{High level flow of the proposed framework. Clustering of weights into weight vectors pool, any fine tuning, activation bitwidth selection is done offline. At inference time, the processor only stores the weight pool dot product results and indices to weight vectors used in the network.}
    \label{fig:high_level_framework}
\end{figure}

%\subsection{Overview of the framework}
Figure \ref{fig:high_level_framework} shows the high-level flow of the proposed framework, which is split into two parts. The left block shows the compression part, where the input is a pretrained CNN. The corresponding weight pool and weight indices (original weights are converted to indices of the weight pool) are generated and the pretrained CNN is hence compressed. Analysis of minimum activation bitwidth of the compressed CNN is carried out afterwards. Finally, the dot product lookup table is generated from the weight pool, and loaded into microcontrollers' flash memory along with weight indices and precision information. The compression part is entirely executed on the host side and the generated weight pool CNN is sent to the microcontroller. 

The second part is CNN inference acceleration, which is executed on the microcontroller. At this stage, the original CNN has already been compressed and transformed into weight pool CNN, and the activation bitwidth has been determined. The framework uses a bit-serial lookup table based algorithm to accelerate the inference of weight pool CNNs, and is able to further improve runtime by reducing the activation bitwidth. 
The rest of this section describes each of these steps in detail.

\begin{figure}[h]
    \centering
    \includegraphics[width=0.8\columnwidth]{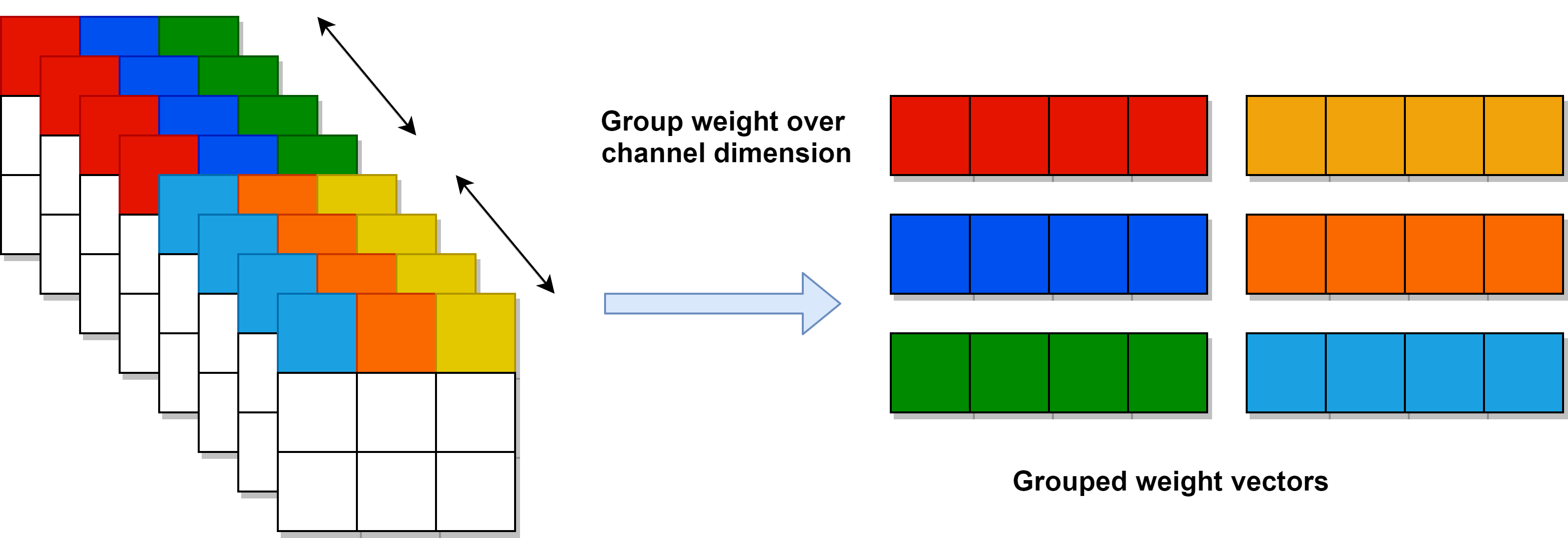}
    \caption{Visualization of z-dimension weight grouping. This example shows a $8\times3\times3$ filter with weight vector size of 4. The weights are grouped in the channel dimension and same color represent weights in a single group. After the z-dimension grouping, 18 $4\times1\times1$ weight vectors (6 are shown in the figure) are generated for the given filter.}
    \label{fig:zdimvisualization}
\end{figure}

Weight-pool networks achieve compression by sharing a fixed pool of weight vectors among all the layers of a network, so that the network only needs to store indices of the weight pool, plus the weight pool itself. %For CNNs, an intuitive way to share the weights is sharing the 2D convolution kernel, since many trained 2D kernels may share similar properties, e.g., edge detectors. \cite{son2018clustering} use k-means clustering algorithm to cluster the $3\times 3$ kernels of a pretrained network and uses the cluster centers as the weight pool and share them among the entire network. However, this approach requires an additional scaling coefficient for every 2D kernel to get satisfactory accuracy, which reduces the compression ratio by $2\times$. 
\begin{figure}[h]
    \centering
    \includegraphics[width=0.6\columnwidth]{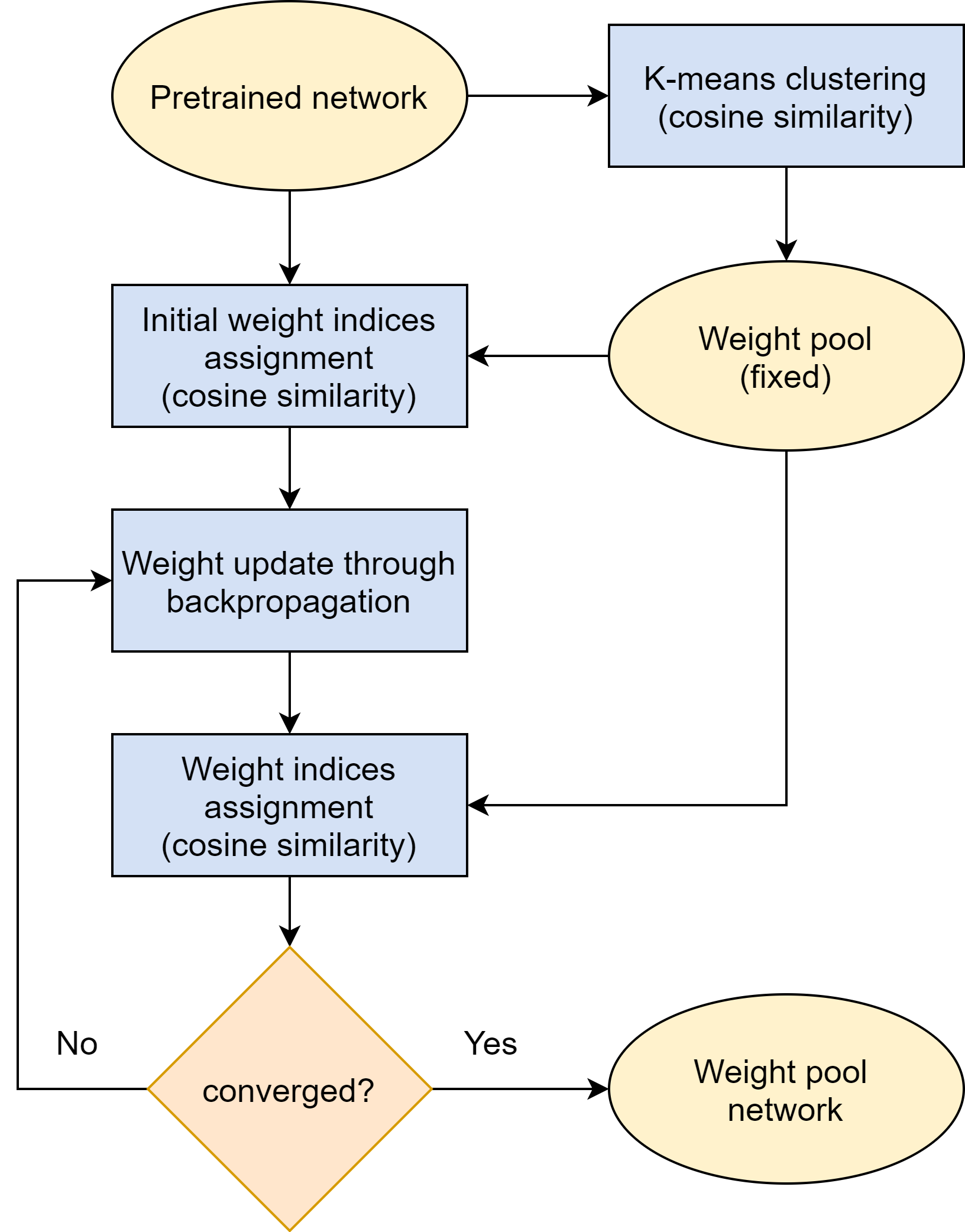}
    \caption{Overall flow of generating a weight pool network from a pretrained network.}
    \label{fig:weightpoolflow}
\end{figure}
In this work we use a weight sharing pipeline similar to  \cite{son2018clustering} to generate weight pool CNNs, but instead of clustering 2D convolutional kernels, we apply the clustering algorithm along the z-dimension of a 3D filter (clustering across the filter channels) as shown in Figure \ref{fig:zdimvisualization}. Figure \ref{fig:weightpoolflow}  shows the proposed training pipeline. The pretrained weights are grouped into $1\times 8$ weight vectors along the channel dimension and clustered using K-means clustering (with a cosine distance metric to avoid scaling dependence). After the  clustered weight pool is generated, the original CNN's weights are converted to the indices of the weight vectors in the weight pool. The network is retrained to fine-tune the weight indices assignment (with a fixed weight pool) and fully connected layer's weights. The backward pass updates the network weights and the forward  pass reassigns indices to the nearest weight pool vector. Weight pool network may be further fine-tuned, if needed, for reduced activation bitwidth.
%To maximize compression ratio and improve performance and flexibility, we made some different design choices compared to \cite{son2018clustering}, including z-dimension weight grouping and using cosine similarity as the distance metric. 

%\paragraph{Z-dimension Weight Sharing}
%Although clustering 2D convolution kernels for CNNs seems more intuitive, grouping and clustering weights in the channel dimension is a better option in terms of compression ratio. Clustering 2D kernels works fine for $3\times3$ kernels, but it requires an scaling coefficient per 2D kernel to get decent accuracy and does not lend itself to other kernel sizes. In the proposed framework we group and cluster weights in z-dimension to generate the weight pool as shown in Figure \ref{fig:zdimvisualization}.

\begin{figure}[h]
    \centering
    \includegraphics[width=1.0\columnwidth]{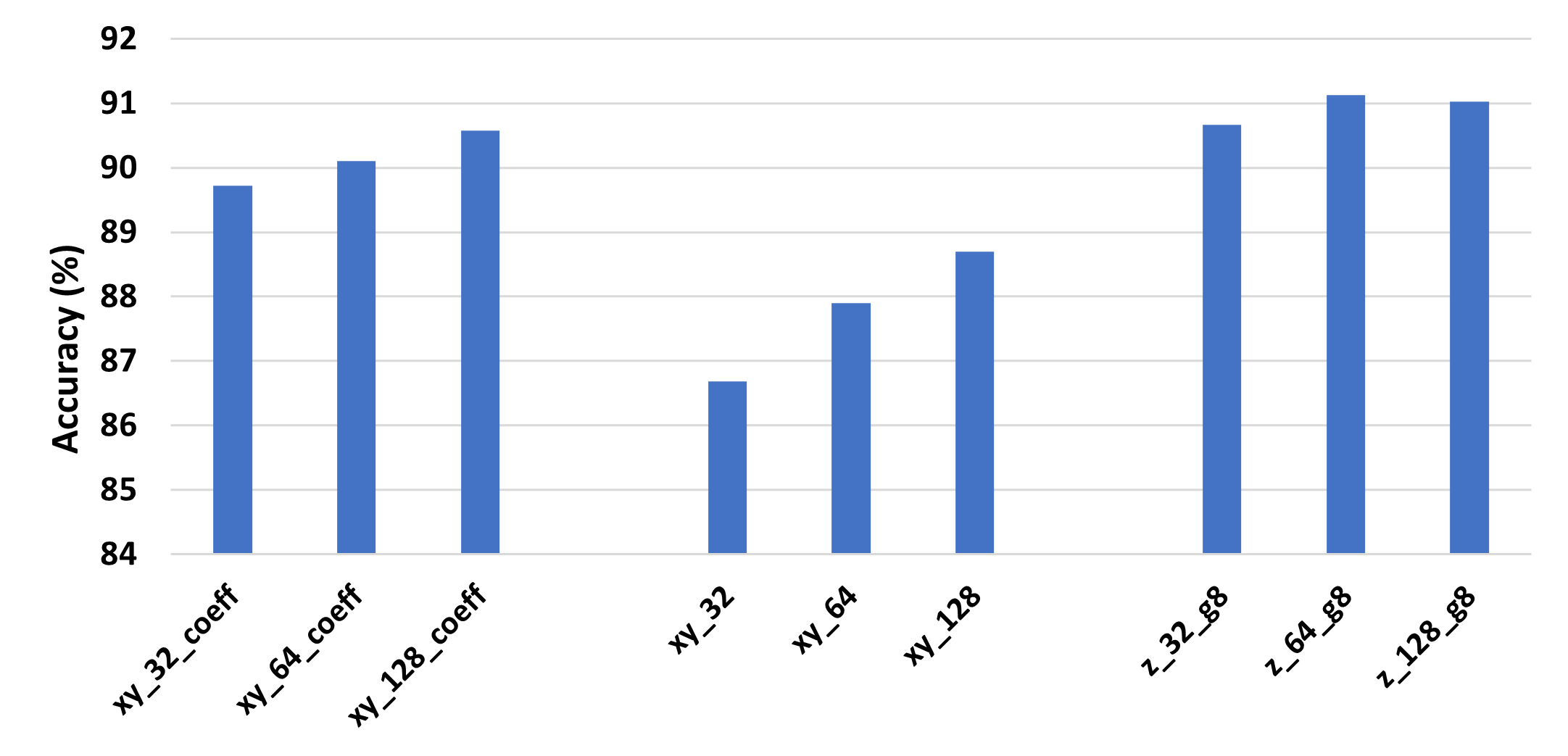}
    \caption{Accuracy of weight pool ResNet-14 with different setups, on CIFAR-10 dataset. For weight pool with $3\times 3$ kernels, its setups are denoted by xy\_n\_(coeff), where n means the weight pool size (how many weight vectors in the weight pool) and coeff means the version with scaling coefficients. For z-dimension weight pool, the setups are denoted by z\_n\_g8, where n is the weight pool size and g8 means the weight vector size (group size) is 8. The original accuracy is 92.26\%.}
    \label{fig:acc_xyz_groupsize}
\end{figure}

To show the effectiveness of z-dimension weight pool and determine the optimal pool size, we benchmark $3\times 3$ kernel weight pool (xy-dimension weight pool) with and without scaling coefficient, as well as the proposed z-dimension weight pool using ResNet-14 (modified ResNet-18 \cite{resnet} with last block truncated) on the CIFAR-10 dataset. For each setup three weight pool size are tested. The result is shown in Figure \ref{fig:acc_xyz_groupsize}. For all three weight pool sizes, z-dimension weight pool performs slightly better than xy-dimension weight pool with coefficients and significantly better than xy-dimension weight pool without scaling coefficients. Regarding the pool size, 64 is enough for this network and 32 also achieves a decent result.

\begin{table}[t]
\begin{tabularx}{\linewidth}{X|l|l|l}
\toprule
Group size    & 4     & 8     & 16    \\ 
Accuracy (\%) & 91.22 & 91.13 & 87.96 \\ \bottomrule
\end{tabularx}
\caption{Accuracy of z-dimension weight pool with different group size. The network is ResNet-14 and dataset is CIFAR-10. Original network accuracy is 92.26\%.}
\label{tab:groupsize}
\end{table}

Table \ref{tab:groupsize} shows the accuracy results of different group size (weight vector size) for z-dimension weight pool on ResNet-14. Clearly, group size of 8 achieves a good balance between compression ratio and network accuracy. We choose 8 as the default group size so the weight pool contains multiple $1\times 8$ weight vectors. Compared to clustering $3\times 3$ kernels, clustering along z-dimension has few advantages:
\begin{itemize}
    \item It achieves same or better network performance (accuracy) without the additional scaling coefficient as used in \cite{son2018clustering}, which improves the compression ratio from $4.5\times $ (clustering $3\times 3$ kernels) to $8\times $ over an 8-bit network. 
    \item It is more flexible. It can fit networks with arbitrary kernel sizes including $1\times 1$ kernels, and can pool  fully connected layers as well. %We indeed compress fully connected layers in the network as well using the same weight pool. 
 %   \item It provides trade-off between compression ratio and network performance by adjusting the group size. 
\end{itemize}

Grouping weights along z-dimension for layers with depth less than 8 (e.g., typical input layers in image CNNs) incurs underutilization. In most, if not all popular CNNs, such reduced depth layers account for a small fraction of storage and compute. Therefore, we choose to keep such layers (usually just the  first layer) uncompressed for better inference accuracy. Another alternative can be grouping all the channels together and zero pad the vector size to 8. 

%However, int the actual implementation we choose to not do this, instead, we keep the first layer uncompressed and implement it using CMSIS library. This is because the percentage of total weight storage consumed by the first layer is almost negligible due to the limited number of channels (input layer also tends to have fewest filters). Similarly, input layer usually is not the most time-consuming layer to execute even though it has the largest activation size. The advantage of not compressing and accelerating the input layer is higher inference accuracy since the input layer typically has high importance. 

%\paragraph{Optimized Distance Metric}
%To make the weight pool generalize better and avoid the extra coefficient for scaling magnitudes, we use cosine similarity as the distance metric for K-means clustering, instead of more commonly used mean square error. Similarity, in the weight indices assignment step we also use cosine similarity as the distance metric to find the neatest weight vector in the weight pool. Doing so the impact of magnitude is eliminated, hence the weight pool generalizes better and achieves better results. The design space exploration of weight pool networks and related accuracy results are reported in section \ref{sec: accuracy}. 

\subsection{Lookup Table Based Bit-serial Computation}
As introduced in section \ref{moitvation}, lookup tables can be used to accelerate convolutions by looking up the vector dot product results directly from memory, instead of computing them. Lookup table offers a trade-off between space complexity and time complexity, and can improve runtime when the memory is large enough and fast enough. However, for dot product operations, the size of lookup table can be huge. Consider the the dot product between two 8-element vectors with 8-bit precision, the total number of entries required for the lookup table is ${{2^8}^2}^8 = 3.40\times 10^{38}$. Clearly, such lookup table implementation is not feasible unless the lookup table size can be massively shrunk. 

The huge lookup table size is partly due to both inputs have no restriction on their values, leading to 65536 total input combinations for a simple two-input multiplication. However, this is not the case for weigh-pool networks. Unlike normal neural networks where inputs and weights can be any possible values, weights are fixed for weight pool networks, meaning a single 8-bit multiplication only requires 256 lookup table entries. The lookup table size for the aforementioned 8-element dot product operation with weight fixed is $1.84\times 10^{19}$ entries, which is significantly smaller than $3.40\times 10^{38}$, but still impractical. 
\begin{figure}[h]
    \centering
    \includegraphics[width=1.0\columnwidth]{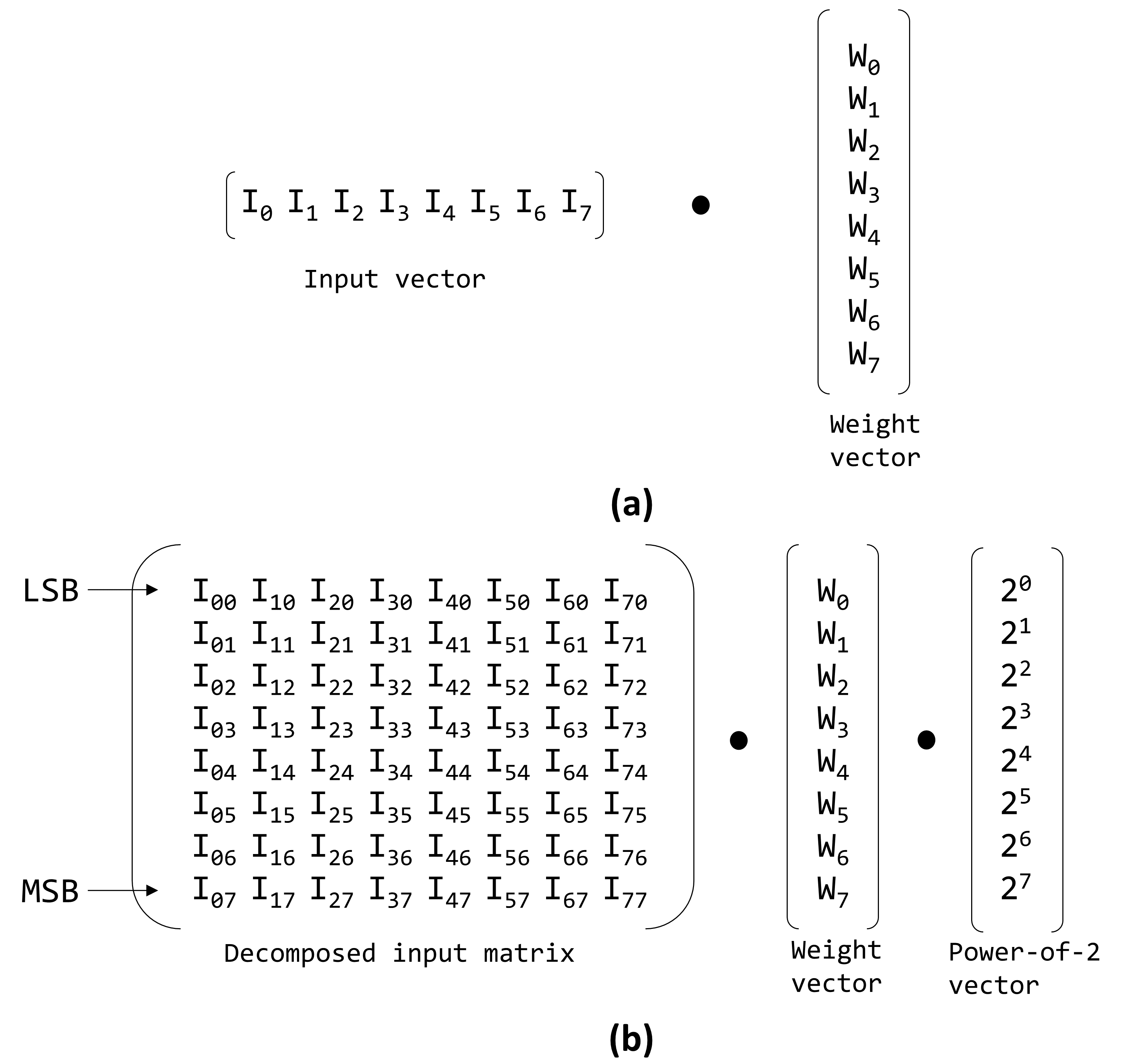}
    \caption{Visualization of the bit decomposition step. (a): The original 8-element dot product between input and weight vectors. (b): The original dot product is transformed into matrix-vector multiplication followed by dot product after bit decomposition. $I_{mn}$ means the $n^{th}$ bit (starting from LSB) of the $m^{th}$ element. The original input vector is decomposed into a $8\times 8$ matrix with each element representing a single bit. Each column represents all the bits of a input value while each row represents a unique bit position of all input values. The weight vector is kept same and is multiplied with all the bit positions of input. The results of the matrix-vector multiplication should be the dot product of input and weight vector at every input bit position. The result is then multiplied with the power-of-two vector which represents bit weights to generate the final dot product result.}
    \label{fig:decompostion}
\end{figure}

To further reduce the lookup table size and support bit-serial multiplication, a key step in our proposed method is \textit{bit decomposition}. For an N-element dot product between input and weight vector (both M bits), the dot product between input (activation) vector and weight vector can be calculated as:
\begin{equation} \label{eq:base_mac}
    \vec{a} \cdot \vec{w} =  \sum_{i=0}^{N-1}a_{i} \times w_{i}
\end{equation}
Where $a_{i}$ and $w_{i}$ are the i-th elements of vectors $\vec{a}$ and $\vec{w}$ respectively and $N$ is the width of the dot product. The input element $a_{i}$ can be decomposed as: 
\begin{equation} \label{eq:wgt_bin}
    a_{i} = \sum_{j=0}^{M-1}2^{j} \times a_{i}[j]
\end{equation}
Where $a_{i}[j]$ is the j-th bit (from LSB) of activation $a_{i}$, and $M$ is the bitwidth of the activation. Hence each input element is decomposed into M binary values each representing a single bit, and the input vector is hence decomposed into an $M\times N$ matrix where each row represents a bit position.
Each time one row (one bit position) of the input matrix is multiplied with the weight vector by looking up the correct dot product result, and then the result is multiplied with the corresponding bit weight. This step is repeated $M$ times until all the bits are processed and all the results are accumulated to calculate the final result. Doing so, the dot product is effectively calculated in a bit-serial way, and it takes $M$ iterations to compute the original dot product. Figure \ref{fig:decompostion} visualizes the decomposition process using the 8-element 8-bit dot product example.

\subsection{Lookup Table Bitwidth and Weight Pool Storage} By decompose the input vector, the lookup table only needs to store the results of the dot product between $N$ 1-bit input elements and $N$ fixed weight elements. The required lookup table size is thus reduced to $2^N$ entries, which is 256 entries for the 8-element dot product example. Assuming 64 fixed weight vectors are needed for a weight pool network (we will show later 64 is enough for most cases), and the results are stored in 8-bit precision, the total lookup table storage for the entire network is just 16 kB. Since the lookup table needs to be stored in memory, this storage overhead should be considered when calculating the overall compression ratio of weight pool networks. Besides the activation/weight vector length $N$, We also denote the lookup table bitwidth by $B_l$ and the size of weight pool by $S$, the formula for lookup table storage in bits is:
\begin{equation}
    Storage_{LUT} = 2^{N}\times S \times B_l
\end{equation}
For a network with $W$ total weight parameters and weight bitwidth of $B_w$, the total network storage in bits is $W \times B_w$. Assuming all the weights of the network are compressed by the weight pool method, the maximum compression ration that can be achieved is:
\begin{equation}
    CR = \frac{W \times B_w}{(\frac{W}{N}\times log_2 S + 2^{N}\times S \times B_l)}
\end{equation}
, where the term $\frac{W}{N}\times log_2 S$ is the weight index storage. $log_2 S$ is the minimum bitwidth required for the weight index, but in actual implementation it may be more efficient to use 8 or 16 bits. 
%The 16 kB storage requirement means it can be put into almost any embedded processor while still leaving enough storage space for compressed weight pool networks. For a moderately-sized CNN, the number of parameters can easily be over a million. The weight indices storage requirement after weight pool compression can still be one hundred to several hundred kB. The 16 kB weight pool storage will not significantly impact the overall compression ratio for most middle to large networks. %Table \ref{fig:relation} summarizes the relationship between weight pool CNN, lookup table and bit-serial computation, and why they are combined together in the proposed framework.
%\begin{figure}[h]
%    \centering
%    \includegraphics[width=0.8\columnwidth]{figures/relationship.png}
%    \caption{Relationship between weight pool compression, lookup table and bit-serial computation in our proposed framework. For each block, the yellow region indicates the method, the red region summarizes the issue related with this method. The green region of the final block summarizes the benefits of the proposed method.}
 %   \label{fig:relation}
%\end{figure}

Interestingly, the weight bitwidth of weight pool networks can be arbitrary since the weights are not explicitly stored. The entire weight pool is converted to a lookup table and the dot product results are stored instead of weights. In this case, the lookup table bitwidth matters, as it determines how much memory space is required for storing the lookup table, as well as the inference accuracy of the network. Storing the lookup table at low bitwidth essentially reduces bitwidth (precision and/or range) of dot-product partial sums and may compromise the inference accuracy. We experimentally show that 8-bit lookup table precision is good enough for most cases. The full results are shown in \ref{sec: accuracy}.

\subsection{Activation Bitwidth and Weight-Pool Network Runtime}
In terms of theoretical runtime performance, for the 8-element 8-bit dot product example, the proposed method requires 8 iterations to loop over bit positions and each iteration contains two memory loads (input and result), one shift and one accumulate operation. The weight indices are same for all the bits and hence can be shared. Normal convolution also requires 8 iterations to loop over individual vector elements and each iteration requires two memory loads (activation and weight), one multiply and one accumulate. This analysis shows that our proposed method has almost identical theoretical runtime compared to 8-bit baseline without considering overheads and optimizations. This is a promising result since the proposed method can have better runtime than the baseline by simply reducing the activation bitwidth below 8 bits. We will show that with various reuse and optimizations, our proposed method has better runtime even at 8-bit activation bitwidth compared to the 8-bit baseline using ARM's CMSIS library \cite{lai2018cmsis}.

\section{Weight Pool Implementation: Overheads and Optimizations}
\label{optimization}
There are many runtime overheads associated with software bit-serial processing and weight sharing. Here we discuss these overheads and the corresponding optimizations to overcome them. 
\subsection{Bit Unpacking Overheads and Optimized Dataflow}
%\paragraph{Bit unpacking}
For software sub-byte precision computation, bit unpacking causes significant runtime overhead since processors typically are byte addressable. For our bit-serial lookup method, the bit decomposition step needs to unpack each element of the input vector into individual bits, and the same bit position of different input elements (rows of the decomposed input matrix in Figure \ref{fig:decompostion}) should be grouped together for lookup table computation. Doing this in software requires iterating over all the input elements and for each input element there is an inner loop to extract all the bits. For the 8-element, 8-bit dot product example, 64 iterations are required for a single dot product, while only 8 iterations are required for the actual computation. Implementing bit unpacking for every dot product can significantly slow down the runtime, making it roughly $9\times$ slower than baseline hence negating any potential speedup by reducing the activation bitwidth. 

%\paragraph{Optimized dataflow}
 To address the bit unpacking overhead, we utilize input reuse in our dataflow so that the bit unpacking step (activation vector decomposition) can be shared. For CNNs, same input can be reused for all the filters of a layer, so that the bit unpacking overhead per result lookup is reduced by a factor equal to the number of total filters in a layer. To implement input reuse and share the bit unpacking overhead, we order the loops such that the filter lookup is inside the loops over input channels and filter x, y dimensions. The activation vector decomposition (bit unpacking) is implemented right before the filter loop, so that the decomposed activation matrix can be reused. Algorithm \ref{code:dataflow} shows the overall flow including the modified loop order. The bit-unpacking step happens at line 7 of Algorithm \ref{code:dataflow}. For a convolution layer with $N$ filters, the time spent on bit unpacking is reduced by a factor of $N$ and is significantly less than the time spent on result lookup for most layers. 

\begin{algorithm}[htbp]
\caption{The simplified algorithm flow of the bit-serial lookup table implementation. Number of input channel group is number of total input channels divided by weight vector size.}
\label{code:dataflow}
 \begin{algorithmic}[1] 
    \FOR{loop over batch} 
        \FOR{loop over output x-dimension}
            \FOR{loop over output y-dimension}
                \FOR{loop over kernel x-dimension}
                    \FOR{loop over kernel y-dimension}
                        \FOR{loop over input channel groups}
                            \STATE \textbf{Activation vector decomposition (bit unpacking}
                            \STATE \textbf{Lookup table caching (flash to ram)}
                            \IF{Precomputation}
                                \FOR{loop over weight pool vectors}
                                    \FOR{loop over activation bits}
                                        \STATE \textbf{Results lookup}
                                        \STATE \textbf{Shift and accumulate}
                                    \ENDFOR
                                    \STATE \textbf{Store results in RAM}
                                \ENDFOR
                                \FOR{loop over filters}
                                    \STATE \textbf{Precomputed results lookup}
                                \ENDFOR
                            \ELSE
                                \FOR{loop over filters}
                                    \FOR{loop over activation bits}
                                        \STATE \textbf{Result lookup}
                                        \STATE \textbf{Shift and accumulate}
                                    \ENDFOR
                                \ENDFOR
                            \ENDIF
                        \ENDFOR
                    \ENDFOR
                \ENDFOR
            \ENDFOR
        \ENDFOR
    \ENDFOR
\end{algorithmic}000000
\end{algorithm}

\subsection{Memory Latency and Lookup Table Caching}\label{lutcaching}
%\paragraph{Flash access latency}
In a typical microcontroller, flash memory is used as the main storage and SRAM is used for holding variables during computation. Flash memory has more storage space than SRAM but operates slower. However, due to SRAM's limited size (typically 16-128 kB), it can only be used to hold activations and some temporary variables. The network weights are normally stored in flash memory (size ranges from 128 kB - 2 MB), and during computation the weights are loaded from the slower flash memory. For weight pool networks, the lookup table size is typically 8-32 kB, which is similar to the SRAM size of some small microcontrollers. For such really tiny, low-cost processors, the lookup table cannot fit in SRAM and need to be stored in flash, hence the result lookup latency will be higher and hurt runtime. 

%\paragraph{Lookup table caching}
To improve the result lookup latency, we cache the active part of the lookup table in SRAM. Before explaining what is the active part of a lookup table, we first discuss how data can be arranged inside a lookup table. The lookup table of the proposed method contains the dot product results between all weight vectors and all possible input (activation) bit vectors. There are two ways to order the lookup table contents when storing them in memory, one is weight oriented order and the other is input oriented order. Visualization of the two lookup table orders are shown in the appendix. Assume the total number of weight vectors in the weight pool is $S$ and the activation bitwidth is $M$. For weight oriented order, the lookup table can be split into $S$ smaller concatenated lookup tables, each containing the results of all possible inputs related to a single weight vector. For input oriented order, the lookup table consists of $2^M$ smaller lookup tables and each of them contains the results of one input with all weight vectors. %The weight oriented order may seem more straightforward compared to input oriented order since the lookup table represents the weight pool. 
Input oriented order is more compatible with input reuse dataflow since a few blocks (results corresponding to the bit-vectors generated by the input matrix decomposition) of the lookup table is repeatedly accessed in the filter loop, with other blocks of the lookup table stay idle. We utilize this property and cache the active blocks of the lookup table from flash to SRAM during computation. We use input oriented lookup table in our implementation to reduce the flash access overhead and improve runtime.

In our implementation, the dataflow is configured to boost input reuse, and lookup table accesses can also benefit from this dataflow by caching the lookup table in SRAM. In our input reuse dataflow, after activation decomposition the activation vector is multiplied with corresponding weight vectors for all filters. In this case, only a portion of the lookup table related to the generated activation vectors will be used inside the filter loop. Still considering 8-bit activation bitwidth and weight pool size of 64. After the bit decomposition step, 8 activation bit vectors are generated. For the input oriented lookup table, only 8 blocks of the original lookup table each with 64 entries (weight pool size) that corresponds to the activation bit vectors will be actively used in the filter loop. The total size of active lookup table is just 512 bytes, which is small enough to fit into most microcontroller SRAMs.

Hence, as shown in line 8 of Algorithm \ref{code:dataflow}, before entering the filter loop, we load the active portion of the lookup table from flash and cache them in SRAM. Figure \ref{fig:lutprefetch} visualizes the lookup table caching process. The overhead of this lookup table caching step is again compensated by sharing it across all the filters. Doing so in the innermost loop the lookup table results will be loaded from SRAM instead of flash, therefore the overall runtime can be improved.
\begin{figure}[h]
    \centering
    \includegraphics[width=1.0\columnwidth]{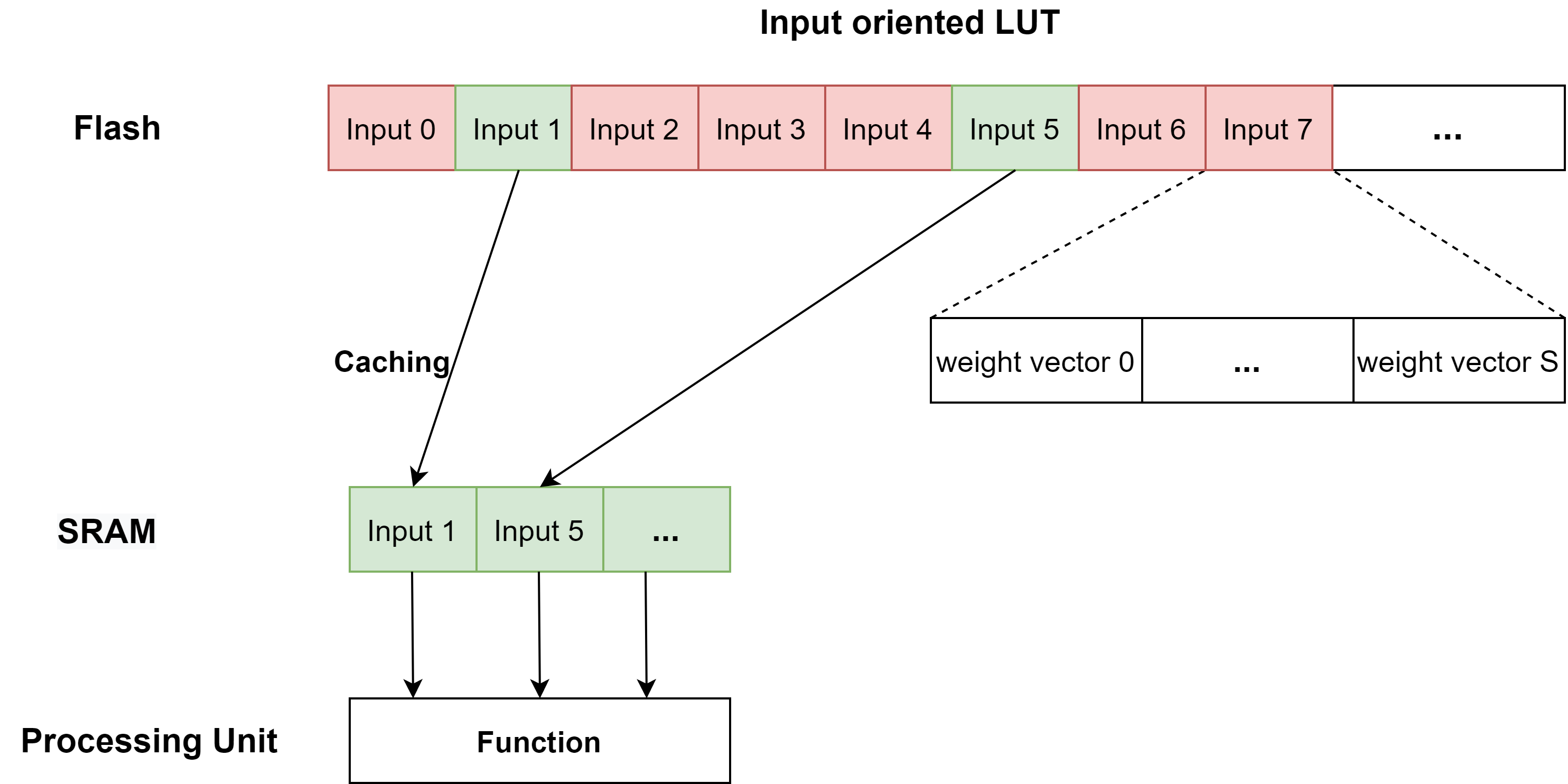}
    \caption{Visualization of lookup table caching. Green blocks represent active lookup table regions corresponding to the input vectors that are shared across filters. Red blocks represent inactive lookup table region. Active regions are cached into SRAM before the filter loop and the function only accesses lookup table results from SRAM. }
    \label{fig:lutprefetch}
\end{figure}

To validate the analysis, we benchmark the lookup table caching optimization against the implementation without lookup table caching (everything else is same) on individual layers with different number of filters. The results are shown in Figure \ref{fig:lutprefetch} (orange bars). The lookup table caching version outperforms baseline for all 4 layer configurations, and the speedup scales with number of filters in the layer (due to better reuse). While lookup table caching only marginally improves runtime for layer with 32 filters, it achieves more than $1.4 \times$ speedup for layer with 192 filters.

\subsection{Weight Pool Computation Reuse Through Precomputation}
The main property of weight pool networks is that a small pool of weight vectors are shared across the entire network. We have shown that using a pool of 32 or 64 8-element weight vectors are enough for maintaining the accuracy, and such pool sizes are often smaller than the number of filters of a large convolution layer, which can be more than 256. The relative small pool size offers computation reuse opportunities on large convolution layers to further improve runtime of weight pool networks.

A property of CNNs is that the same input vector can be reused for all the filters of a convolution layer. For weight pool networks, weights are selected from a group of weight vectors and the total number of distinct weight vectors is the pool size (32 or 64). If a convolution layer has more filters than the pool size, an input vector will inevitably multiply with some weight vectors multiple times when looping over filters. In other words, for a weight pool network, the maximum number of unique dot products that needs to be computed for a given input vector is the weight pool size, regardless the actual number of filters in that layer. To avoid unnecessary computation for large convolution layers, precomputation can be used to only compute the necessary dot products between inputs and weights and store them in another lookup table, hence repeated (bit-serial) computation will be replaced with result lookups. 
Another way to avoid repeated computation is memoization, where the dot product resutls is dynamically memoized during computation (inside the filter loop). We compare and evaluate the two methods (analysis is in appendix) and precomputation performs better. The simplified flow of precomputation is shown in lines 9-16 of Algorithm \ref{code:dataflow}. 

Precomputation should only be used for large convolution layers as its benefits rely on large number of filters (it improves runtime when the number of filters of a layer is larger than the weight pool size). For a given layer, precomputation is used only when the number of filters is \textit{larger} than the pool size. To demonstrate the effectiveness of precomputation, we combine precomputation with lookup table caching and evaluate the speedup against baseline implementation, using the same benchmark in section \ref{lutcaching}. The results in figure \ref{fig:lutprefetch} shows that for layers that have more filters than the weight pool size, precomputation can further improve the runtime of lookup table caching version. For layer with 192 filters, precomputation + lookup table caching achieves $2.45\times$ speedup against baseline implementation and is $1.7\times$ faster than just using lookup table caching. However, for layers with number of filters that smaller or equal to the weight pool size, precomputation hurts runtime. This result support our analysis that for those layers precomputation should not be used. 
\begin{figure}[t]
    \centering
    \includegraphics[width=1.0\columnwidth]{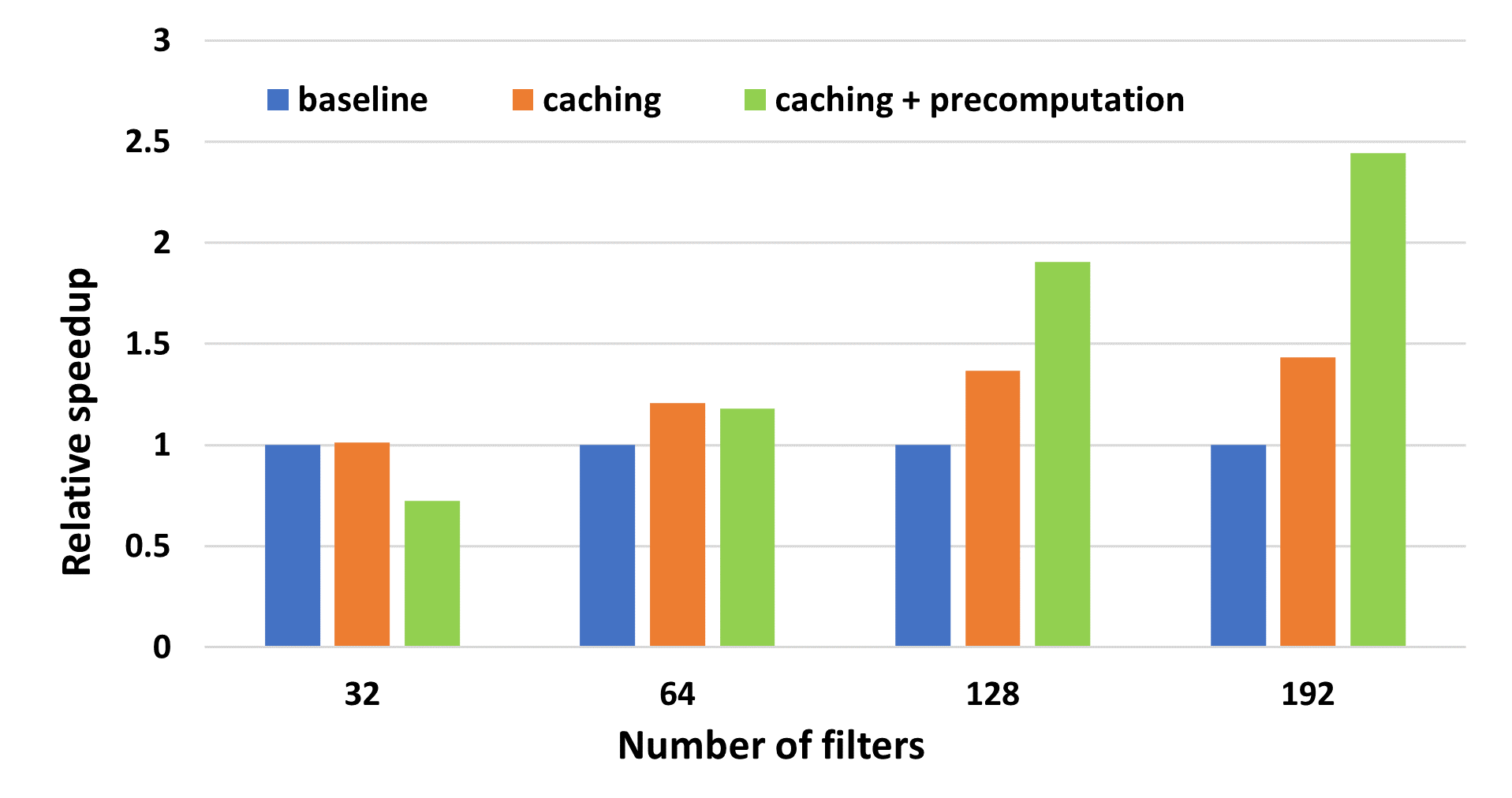}
    \caption{Relative speedup of just lookup table caching (orange) and precomputation + lookup table caching (green) against baseline implementation. Four $3\times 3$ convolution layers with different number of filters are tested. The number of channel is set to be same as number of filters and the input size is $16 \times 16$. Weight pool size is 64.}
    \label{fig:lutprefetch} 
\end{figure}
\vspace{-0.1in}

\begin{table}[b]
\begin{tabularx}{\linewidth}{l|lXXXX}
\hline
Name     & Model  & SRAM (kB) & Flash (kB) & Core & Freq. (MHz) \\
MC-large & F207ZG & 128  & 1024  & CM3   & 120       \\
MC-small & F103RB & 20   & 128   & CM3   & 72        \\ \hline
\end{tabularx}
\caption{STM Nucleo family microcontrollers used for benchmarking. Both use ARM Cortex M3 for the core.}
\label{tab:mcinfo}
\end{table}

\paragraph{Run-time accuracy trade-off.}
Precomputation not only accelerates wide convolution layers, it also offers another way to make trade-off between runtime and accuracy, besides adjusting the activation precision. For a relatively wide network that contains layers wider than 32 filters, the runtime can be improved by reducing the weight pool size. Although we observed that weight pool size of 64 works reasonably well in most cases and set as default size, 32 is also good enough for many cases. The runtime can be improved with tiny drop in accuracy by reducing the weight pool size in such cases.

\section{Evaluation}

\subsection{Experimental Setup}
We evaluate the accuracy and runtime of z-dimension weight pool method on five different networks: TinyConv \cite{lai2018cmsis}, MobileNet-v2 \cite{sandler2018mobilenetv2}, ResNet-10 (ResNet-18 with last two blocks truncated), ResNet-14 (ResNet-18 with last block truncated) and ResNet-s (scaled-down version of ResNet-18 used in \cite{banbury2021mlperf}). We use 2 datasets, CIFAR-10 and Quickdraw-100 (100 classes), and forms 5 network-dataset combinations. All ResNets are tested on CIFAR-10 and MobileNet-v2 and TinyConv are tested on Quickdraw-100. The network structures are adjusted slightly to fit CIFAR-10 and Quickdraw-100. For the weight pool version of MobileNet-v2, only the $1\times 1$ point-wise convolution layers are compressed using weight pool. Depth-wise convolution layers are kept uncompressed since they do not fit our proposed implementation. Theoretically the depth-wise layers can be compressed using xy-dimension weight pool, but it is not necessary - those layers accounts for a very small portion of storage (2.93\%) and runtime.

All the accuracy results are evaluated using PyTorch framework. For network training and retraining, SGD is used as the optimizer with learning rate scheduling, and batch size set to 128.
For runtime results, we use two microcontrollers as shown in Table \ref{tab:mcinfo}. We use ARM Compiler version 6 and runtime is measured using the built-in cycle counter. The frequency is set to maximum frequency for both boards.  

\subsection{Compression Ratio}
\begin{table}[htbp]
\footnotesize
\begin{tabularx}{\linewidth}{l|XXl}
\hline
Network      & Total param & CR   & LUT overhead \\ \hline
TinyConv     & 81600              & 2.32 & 29.8\%       \\
ResNet-s     & 170928             & 4.43 & 29.7\%       \\
ResNet-10    & 665280             & 6.51 & 13.8\%       \\
ResNet-14    & 2729664            & 7.55 & 4.3\%        \\
MobileNet-v2 & 2249792            & 6.22 & 4.5\%            \\ \hline
\end{tabularx}
\caption{Total number of parameters (uncompressed), overall compression ratio (CR) and lookup table overhead of the selected networks. The lookup table overhead is the proportion of lookup table storage to the total network storage after compression.}
\label{tab:compression}
\end{table}

Table \ref{tab:compression} shows the total number of parameters and overall compression ratio of the networks with weight pool size of 64. The lookup table overhead is also shown and is compression limiting only for small networks such as TinyConv. The compression ratio improves as the network size increases, and is close to the theoretical maximum ($8 \times$) for ResNet-14 (and even larger networks). Smaller networks further suffer in compression since the first convolution layer and fully connected layers are not compressed, whose effect is not well amortized. \footnote{Compressing fully connected layer with weight pools improves compression ratio for Resnet-s (TinyConv)  to 4.5(3.1) but at the cost of 0.7\%(2.8\%) in additional accuracy drop. These compression ratios improve further to 5.7 (4.2) if weight pool size of 32 is used albeit, again at 0.5\%-1\% additional accuracy drop. In this work we do not compress them as they do not improve compression for most networks but affect accuracy.} 

\subsection{Accuracy Evaluation}\label{sec: accuracy}
\subsubsection{Weight Pool Size}
We first study impact of weight pool size alone on accuracy without any quantization effects. Table \ref{tab:zdimacc} shows the accuracy of z-dimension weight pool compression with three weight pool sizes without any activation quantization compared to an uncompressed  floating-point baseline.  A {\em weight pool size of 64} ensures little accuracy drop for most networks and is our default for all experiments unless otherwise mentioned. ResNet-s, being already compressed, is tougher to compress without accuracy loss. The results demonstrate the effectiveness of z-dimension weight pool compression, even for already small CNNs like TinyConv and ResNet-s.
%\vspace{-0.1in}
\begin{table}[htbp]
%\scriptsize
\footnotesize
\begin{tabularx}{\linewidth}{l|lXXX}
\hline
Network      & Original  & 32    & 64     & 128     \\ \hline
\multicolumn{5}{c}{CIFAR-10} \\ \hline \hline
ResNet-s     & 85.3 & 82.0 & 83.0 & 84.0 \\
ResNet-10    & 91.0  & 89.3 & 89.8 & 90.1 \\
ResNet-14    & 92.3 & 90.7 & 91.1 & 91.0 \\ \hline
\multicolumn{5}{c}{Quickdraw-100} \\ \hline \hline
TinyConv     & 82.2 & 81.7 & 82.2 & 82.3 \\
MobileNet-v2 & 86.5 & 86.7 & 86.8 & 86.9 \\ \hline
\end{tabularx}
\caption{Accuracy (\%) of z-dimension weight pool with different weight pool sizes on selected network-dataset combinations. Original means original network accuracy and 32/64/128 is the weight pool size. }
\label{tab:zdimacc}
\end{table}
\vspace{-0.2in}
\subsubsection{Lookup Table Bitwidth}
For the proposed bit-serial lookup table implementation, dot product results between decomposed activation bit-vectors and weight vectors are stored in the lookup table, and the bitwidth of the lookup table may affect inference accuracy.

To evaluate the impact of lookup table bitwidth on network accuracy, we simulate the proposed bit-serial lookup implementation using PyTorch. Results in table \ref{tab:lutprec} show that a {\em lookup table bitwidth of 8} loses no accuracy and is default for our experiments unless otherwise mentioned. Furthermore, since most processors are byte-addressable, using a bitwidth smaller than 8 would incur performance overheads albeit delivering a better storage compression for small networks. 

\begin{table}[!htbp]
\footnotesize
\begin{tabularx}{\linewidth}{l|XXXX}
\hline
& \multicolumn{4}{c}{Lookup table bitwidth} \\
Network      & No-LUT  & 16    & 8     & 4     \\ \hline
\multicolumn{5}{c}{CIFAR-10} \\ \hline \hline
ResNet-s     & 83.0 & 83.0 & 82.9 & 82.3 \\
ResNet-10    & 89.6  & 89.9 & 89.9 & 89.4 \\
ResNet-14    & 91.1 & 91.1 & 91.1 & 90.4 \\ \hline
\multicolumn{5}{c}{Quickdraw-100} \\ \hline \hline
TinyConv     & 82.2 & 82.2 & 82.1 & 81.6 \\
MobileNet-v2 & 86.8 & 86.6 & 86.6 & 85.5 \\ \hline
\end{tabularx}
\caption{Inference accuracy (\%) of bit-serial lookup table implementation. No-LUT column shows accuracy that not using lookup table implementation. The activation bitwidth is 8 bit.}
\label{tab:lutprec}
\end{table}

\subsubsection{Activation Bitwidth}
Although activation bitwidth does not affect the storage of a weight pool network, it affects the runtime when the weight pool network is implemented using the proposed bit-serial lookup table approach.  We use an iterative search algorithm to determine the optimal range when quantizing activations. The weight pool size is 64 and lookup table bitwidth is 8 for all cases. 
Table \ref{tab:actretrain} shows that for 8-bit activation bitwidth, almost all networks achieve floating point accuracy (i.e., "64" column in Table \ref{tab:zdimacc}). At 5-bit activation bitwidth, most networks still maintain less than 1\% accuracy drop except for MobileNet-v2 which is quantization-unfriendly \cite{mobilenetquantization1,mobilenetquantization2}. Moreover, for lower bitwidths, the accuracy drop can be compensated by retraining the network with activation quantization.  After retraining, activation bitwidth can go down to 3-4 bit within 1\% accuracy drop for all networks except for MobileNet-v2, which requires 5 bits. 
\begin{table*}[t]
\footnotesize
\begin{tabularx}{\linewidth}{l|XXXXXX|l}
\hline
& \multicolumn{6}{c|}{Activation bitwidth} & Min. bitwidth\\
Network      & 8    & 7    & 6    & 5          & 4          & 3   &$<$ 1\% a.d       \\ \hline
\multicolumn{7}{c}{CIFAR-10} \\ \hline \hline
ResNet-s     & 82.9 & 83.0 & 83.1 & 82.9       & 82.5       & 80.4(80.4)    &4   \\
ResNet-10    & 89.9 & 89.9 & 89.8 & 89.6       & 88.9(89.2) & 84.5(87.8) &4\\
ResNet-14    & 91.1 & 91.1 & 91.0 & 90.8       & 90.6(91.0) & 88.5(90.2) &3\\ \hline
\multicolumn{7}{c}{Quickdraw-100} \\ \hline \hline
TinyConv     & 82.1 & 81.8 & 81.2 & 79.3(82.0) & 69.2(81.2) &36.0(77.4) &4\\
MobileNet-v2 & 86.6 & 86.5 & 86.0 & 83.6 (85.9) & 77.9(84.0) & 36.4(73.0)  &5 \\ \hline
\end{tabularx}
 \caption{Inference accuracy (\%) of weight pool networks with different activation bitwidths. Results in brackets are accuracy after retraining. The last column shows the minimum activation bitwidth with less than 1\% accuracy drop. The lookup table bitwidth is set to 8 bit.}
\label{tab:actretrain}
\end{table*}

\subsection{Runtime Evaluation}
\subsubsection{Impact of Activation Bitwidth}
\begin{figure}[h]
    \centering
    \includegraphics[width=1.0\columnwidth]{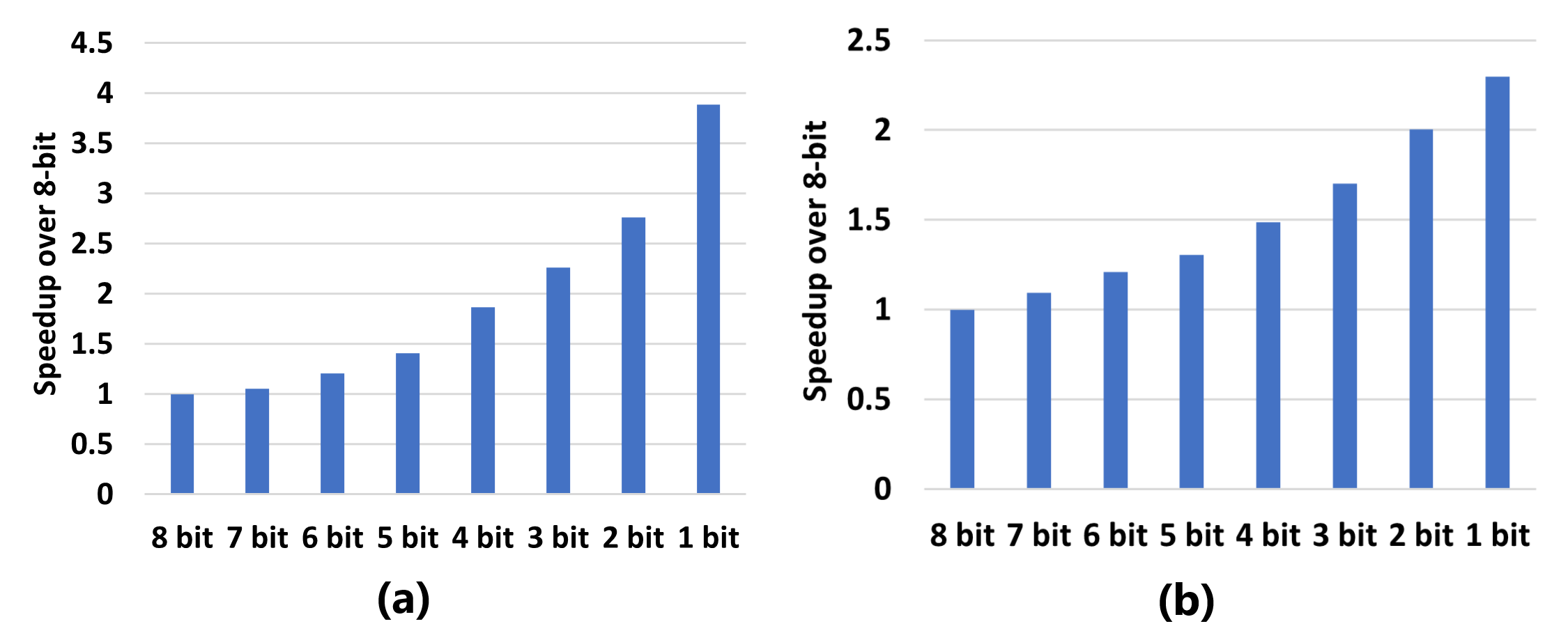}
    \caption{Speedup against 8-bit bit-serial lookup implementation for different activation bitwidths. (a): results without precomputation. (b): results with precomputation. The input size is $16 \times 16$ and number of channels and filters are both 128. Weight pool size is 64. }
    \label{fig:runtimevsprec}
\end{figure}

One of the main contributions of the proposed framework is the support of accelerating runtime by reducing activation bitwidth. We evaluate  the runtime improvement  from a 8-bit baseline on a layer with 128 channels and filters and pool size of 64 in Figure \ref{fig:runtimevsprec}, using MC-large.  Without precomputation, the speedup scales linearly according to activation bitwidth, and is almost $4 \times$ for 1-bit activation (less than the $8 \times$ theoretical speedup due to the fixed bit unpacking overhead). For the precomputation case, as the activation bitwidth reduces, the runtime of bit-serial loop during precomputation reduces, but the runtime for precomputed results lookup does not change and starts to dominate the runtime. However, precomputation already accelerates the runtime significantly so the overall speedup is still better for large layers.

\subsubsection{Full-network Benchmark}
To evaluate the overall runtime performance of the proposed method, we evaluate the full-network runtime performance on both microcontrollers with weight pool size of 32 and 64, and compare with ARM CMSIS implementation whenever possible. The results are shown in Table \ref{tab:fullruntime}. For the minimum activation bitwidth case, the results for 32-vector weight pool are for reference only, since the minimum bitwidth is determined from the results of 64-vector weight pool. %For MC-small, due to limited RAM capacity, we integrate the batchnorm, activation function and pooling into the first layer's convolution function to reduce the output size. 
\vspace{-0.1in}
\begin{table}[htbp]
\footnotesize
\begin{tabularx}{\linewidth}{l|XXXXX}
\hline
Network      & CM.             & 64-8 & 32-8 & 64-m  & 32-m  \\ \hline
\multicolumn{6}{c}{MC-large} \\ \hline \hline
TinyConv     & 1.06              & 0.83 & 0.75 & 0.60 & 0.57 \\
ResNet-s     & 0.60              & 0.49 & 0.43 & 0.31 & 0.28 \\
ResNet-10    & 5.28              & 3.00 & 2.22 & 1.87 & 1.61 \\
ResNet-14    & / & 3.46 & 2.59 & 1.92 & 1.73 \\
MobileNet-v2 & /                  & 3.60     & 3.12     & 3.07     &  2.78    \\ \hline
\multicolumn{6}{c}{MC-small} \\ \hline \hline
TinyConv &      1.95     &  1.49    & 1.33     & 0.99     &  0.89    \\
ResNet-s &       1.24      & 1.07     &  0.89    &  0.63    &  0.55    \\ \hline
\end{tabularx}
\caption{Full-network inference latency (in seconds) with different setup for both microcontrollers. CM. stands for CMSIS implementation, -8 means 8-bit activation precision while -m means minimum activation precision that has less than 1\% accuracy drop that determined in Table \ref{tab:actretrain}. 32 and 64 are the weight pool size. / means the network cannot fit into flash.}
\label{tab:fullruntime}
\end{table}

For all setups, the proposed implementation achieves better runtime than CMSIS and the speedup is better for larger networks. With less than 1\% accuracy drop, the "right bitwidth" weight pools can achieve over $2.8\times$ speedup over CMSIS for medium-sized CNNs like ResNet-10 and  around $2\times$ speedup for smaller CNNs like ResNet-s and TinyConv. There are several factors that make the speedup smaller for small CNNs, including lack of precomputation opportunity, more bit unpacking overhead and bigger impact of not accelerating the first layer. Larger CNNs (ResNet-14, MobileNet-v2) do not fit into the microcontroller memory without the weight pool compression and hence a runtime comparison is not possible.
Overall, the proposed method improves CMSIS runtime on CNNs regardless of network structure and activation bitwidth, and the speedup is larger for large networks. 

\subsection{Comparison with Binarized Networks}
The theoretical compression ratio of a weight pool network is similar to the compression ratio of a binarized networks but with much better accuracy.
\cite{romaszkan20203pxnet} evaluates the implementation of binarized networks on microcontrollers and reports  $2-4\times$ speedup compared CMSIS 8-bit implementations. For comparison, we trained the binarized version of TinyConv and the accuracy for CIFAR-10 is barely 66.9\% as opposed to 81.2\% with weight pools. 

\section{Related Work}

\subsection{Neural Network Weight Sharing}
The concept of weight sharing in neural networks can be dated back to 1992 \cite{softweightsharing1992}, as an approach to simplify neural networks. Recently, weight sharing has been applied to convolution neural networks, by clustering and sharing 2D convolution kernels \cite{son2018clustering,wu2018deepkmeans} for all layers of the network. With tiny or almost no drop in accuracy, weight sharing can significantly compress the parameters of the neural network, which leads to 4.5$\times$ to 36$\times$ reduction in CNN's storage requirement, depending on exact sharing method and baseline precision. 

\subsection{Lookup Table Based Vector Multiplication Acceleration}
Lookup table is a widely used method to improve runtime by replacing computation with memory lookup. There are many works \cite{deng2019lacc, sutradhar2020ppim, ferreira2021pluto} try to accelerate deep neural networks with lookup tables by memorizing vector multiplication results. However, due to the huge lookup table size (GB+) required for memorizing all possible results of a vector-vector multiplication, all of them are DRAM based in-memory accelerators, hence they are not software solutions.

\subsection{Software Based Convolution Acceleration for Sub-byte Precision}
There are a few software-focused works that develop algorithms to deploy sub-byte neural networks on CPUs. \cite{yu2019tf} utilizes a single multiplication instruction to implement multiple sub-byte multiplications through bit-packing, and is able to show performance improvement for four-bit input and ternary weight network over 16-bit baselines. \cite{cowan2018automating} proposes a software method and corresponding optimizations for CPUs to compute sub-byte precision more efficiently. However, as their method has a time complexity proportional to total number of weight bits times total number of activation bits, moderate speedup over 8-bit baseline can only be demonstrated on very low activation and weight bitwidth (2-3 bits). 
Current software methods for accelerating sub-byte neural networks have limited use cases due to their strict requirements on activation and weight bitwidth. For many applications, quantizing both activation and weight to 2-3 bits can severely impact the learning capability of neural networks. Besides, some versions require advanced instructions that are not available for low-power microcontrollers like ARM Cortex M0 and M3. We do not directly compare against these works as the target applications and platforms are not same and they do not offer arbitrary sub-byte precision acceleration.

\section{Conclusion}
We have proposed the first framework for efficiently deploying weight pool networks on resource-constrained processors, with compression, training and execution  methodologies. The proposed weight pool networks with bit-serial lookup table implementation support and accelerate arbitrary sub-byte precision execution, and can achieve up to  $2.8 \times$ speedup and up to $7.5\times$ compression compared to 8-bit networks, with less than $1\%$ drop in accuracy. The proposed framework is more efficient on large networks, both in terms of compression and speedup, therefore is suitable for deploying large neural networks on small microcontrollers.  We are able to fit and accelerate relatively large CNNs like MobileNet-v2 on a microcontroller with 1MB Flash memory, which otherwise will not fit in the processor memory.\footnote{ We plan to make the framework available publicly under the ACM artifact submission guidelines in the future. }

\section*{Acknowledgements}
We thank the ONR for the funding support of this project.

%\clearpage
\bibliography{egbib}
\bibliographystyle{mlsys2022}

\clearpage
\end{document}